# Beyond Fine-Tuning: Evaluating Prompt Engineering and Prompt Tuning for Roman Urdu Hate Speech Detection

**Toneema Zubair [1], Muhammad Junaid Asif [2,3*], Faisal Kamiran [1], Hafiz Hassan Saeed [1], Rana Fayyaz Ahmad [2]**

[1] Department of Computer Sciences, Information Technology University (ITU), Lahore 54000, Pakistan
[2] Natural Language Processing (NLP) Lab, Artificial Intelligence Technology Centre (AITeC), Islamabad 44000, Pakistan
[3] Faculty of IT and Computer Sciences, University of Central Punjab (UCP), Lahore (54000), Pakistan

*junaid.asif@ncp.edu.pk

***Abstract*— Low Resource Languages (LRLs) pose significant challenges for hate speech detection because of a scarcity of readily available annotated data, language variation, non-standard spelling, and context-dependent ambiguity. Beyond the problem of informal conventions of writing in Roman Urdu, there is also the issue of code-mixing. After working on the previous project about zero-shot learning and parameter-efficient fine-tuning (PEFT), this research systematically compares prompt engineering and prompt tuning for Roman Urdu hate speech detection. To study the effect of in-context examples, we will compare the zero-shot and few-shot prompting approaches. To understand the advantage of structured prompting and learnable prompt representations, we will conduct studies of manual vs. mixed prompt-based adaptation strategies. Experimentations are carried out on the PURUT dataset (~72,771 social media comments) with GPT-3.5, LLaMA-3, Mistral-7B, and multilingual BERT in several settings. Experimental results show that few-shot prompting always outperforms zero-shot prompting, with the Mistral-7B model scoring 0.67 F1 compared to 0.56 of the zero-shot models. Based on this observation, prompt-based adaptation is proposed to introduce additional improvements, with the proposed mixed prompt configuration able to yield a F1-score up to 0.87 which further surpasses the performance of regular manual prompts. Despite the larger size of the model, its performance trade-off between inference-based prompting, prompt optimization, and parameter-efficient model adaptation is compared to the previously reported PEFT-based F1-score, which is ~0.93. Overall, the study demonstrates the potential of carefully designed and optimized prompting strategies for improving hate speech detection in linguistically complex, low-resource environments while providing practical insights into selecting suitable adaptation approaches.***

## I. Introduction

The rapid expansion of social media platforms has transformed how people communicate, share opinions, and engage in public discourse. While platforms such as Twitter (now X), Facebook, and YouTube have increased opportunities for information sharing and public interaction, they have also facilitated the dissemination of harmful content, including hate speech, offensive language, and abusive expressions. Hate speech poses significant psychological and societal risks by contributing to discrimination, polarization, and hostility toward individuals and communities. Consequently, automatic hate speech detection has emerged as an important research area within natural language processing (NLP), supporting the development of safer and more inclusive online environments.

Over the past decade, hate speech detection has advanced significantly through machine learning, deep learning, and transformer-based models such as BERT, which leverage contextual embeddings and self-attention mechanisms. However, most studies have focused on high-resource languages like English, where large annotated datasets are available [1]. In contrast, low-resource languages remain underexplored, and models trained on high-resource languages often fail to generalize effectively.

Roman Urdu presents particular challenges for hate speech detection because of its non-standard spelling, phonetic variations, informal expressions, and frequent code-mixing. These characteristics complicate contextual understanding and classification. Furthermore, limited linguistic resources and imbalanced datasets affect model performance. To address these challenges, Saeed et al. [2] introduced the PURUTT dataset, comprising 72,771 toxic and non-toxic social media comments, providing a benchmark for Roman Urdu toxicity classification.

Recent advancements in large language models (LLMs), including GPT, LLaMA, and Mistral, have introduced prompt-based learning as an alternative to conventional model fine-tuning. Prompt-based learning can be broadly divided into **prompt engineering** and **prompt tuning**. Prompt engineering enables models to perform downstream tasks through natural language instructions without updating model parameters. In particular, **zero-shot prompting** relies solely on task instructions, whereas **few-shot prompting** incorporates a limited number of labeled examples to improve contextual understanding and classification performance [3]. These approaches are particularly relevant to low-resource NLP applications where extensive model training may be impractical.

Beyond prompt engineering, prompt tuning provides an alternative mechanism for adapting pre-trained models through task-specific prompt representations. Lester et al.[4] demonstrated the effectiveness of learnable soft prompts, while Li and Liang [5] introduced prefix tuning, which optimizes continuous prompt representations while keeping the underlying model parameters frozen. Additionally, cloze-based prompting reformulates classification as a masked-token prediction task, allowing models to associate contextual information with predefined class labels [6]. These approaches offer different mechanisms for improving task adaptation while reducing the need for full-model fine-tuning.

Despite their advantages, prompt-based approaches remain sensitive to prompt structure, instruction wording, demonstration selection, and linguistic variability. These limitations are particularly relevant to Roman Urdu, where inconsistent spelling and contextual ambiguity can affect classification reliability. Although parameter-efficient fine-tuning (PEFT), particularly Low-Rank Adaptation (LoRA), provides effective task-specific adaptation [8], the comparative effectiveness of prompt engineering and alternative prompt tuning strategies for Roman Urdu hate speech detection requires further investigation.

Building upon our previously published work comparing zero-shot prompting and PEFT-based adaptation [9], this study focuses on a systematic evaluation of zero-shot versus few-shot prompt engineering and prefix versus cloze prompt formulations for Roman Urdu hate speech detection. Experiments are conducted using the PURUTT dataset and multiple pre-trained language models to investigate the influence of prompt design and adaptation strategies on classification performance. The study evaluates accuracy, precision, recall, and F1-score to identify effective prompting approaches for linguistically complex and low-resource environments. To guide this study, the following research questions are formulated:

- **RQ1:** How does few-shot prompting perform compared to zero-shot prompting in Roman Urdu hate speech detection?
- **RQ2:** How do prefix and cloze prompting strategies affect classification performance?
- **RQ3:** Does mixed prompt tuning outperform conventional manual prompting?
- **RQ4:** How do prompt-based approaches compare with previously reported PEFT results?
- **RQ5:** What are the performance and computational efficiency trade-offs among these approaches?

This paper makes several key contributions to the field of low-resource NLP:

- A comprehensive comparison of prompt engineering, prompt tuning, and parameter-efficient fine-tuning for hate speech detection in Roman Urdu.
- Introduction and evaluation of a mixed prompt tuning approach that combines manual templates with learnable soft tokens.
- Empirical analysis of zero-shot and few-shot prompting strategies in low-resource settings.
- Insights into the strengths and limitations of prompt-based learning compared to PEFT.
- Practical recommendations for selecting appropriate adaptation strategies in resource-constrained environments.

The remainder of this paper is organized as follows. Section II reviews related work, while Section III presents the dataset, methodology, and experimental setup. Section IV reports the experimental results, followed by a discussion of key findings in Section V. Finally, Section VI concludes the study and outlines future research directions.

## II. Related Work

This section presents a comprehensive overview of these approaches, highlighting their strengths, limitations, and applicability in such challenging settings.

### A. *Traditional Machine and Deep Learning Approaches*

Early research in hate speech detection primarily employed traditional machine learning algorithms, including Support Vector Machines (SVM), Logistic Regression, Naïve Bayes, Random Forest, and K-Nearest Neighbors (K-NN). These methods typically relied on handcrafted textual features, such as n-grams, TF-IDF, and lexical characteristics, to identify harmful content [7], [8]. Their simplicity, computational efficiency, and effectiveness in structured classification tasks made them widely adopted. In particular, SVM and Logistic Regression demonstrated competitive performance when combined with appropriate feature engineering, while ensemble methods improved classification robustness. Despite these advantages, traditional approaches have limited capabilities in capturing semantic relationships and contextual information. Their dependence on handcrafted features reduces adaptability across languages and domains. These limitations are particularly significant for Roman Urdu, where inconsistent spelling, phonetic variations, code-mixing, and informal expressions complicate text classification[2]. Consequently, traditional classifiers often struggle to identify implicit hate speech, sarcasm, and context-dependent expressions. To overcome these limitations, researchers introduced deep learning architectures, including Convolutional Neural Networks (CNNs), Long Short-Term Memory (LSTM), and Bidirectional LSTM (Bi-LSTM). Unlike traditional methods, these models automatically learn hierarchical textual representations, reducing the need for manual feature engineering. CNNs effectively capture local linguistic patterns, while LSTM-based architectures model sequential dependencies and contextual relationships[9], [10].

Furthermore, attention mechanisms improve classification by emphasizing relevant words and expressions. Deep learning methods have demonstrated improved semantic understanding and greater adaptability to noisy textual data. However, their effectiveness often depends on the availability of sufficiently large annotated datasets. For low-resource languages such as Roman Urdu, limited labelled data can lead to overfitting and reduced generalization. Additionally, model training may require considerable computational resources, restricting practical deployment in resource-constrained environments [10]. These challenges have encouraged the adoption of transformer-based architectures and prompt-based learning strategies that leverage pre-trained language representations for more effective low-resource hate speech detection.

### B. *Transformer-Based Models and LLMs for HSD:*

Transformer-based architectures have significantly advanced natural language processing (NLP) through self-attention mechanisms that capture long-range dependencies and contextual relationships. Among these architectures, BERT has demonstrated strong performance in text classification by generating bidirectional contextual representations, enabling more effective interpretation of complex linguistic patterns and ambiguous expressions[11]. These capabilities have made transformer-based models particularly useful for hate speech detection, where understanding contextual meaning is essential. Despite their effectiveness, conventional transformer-based models often require task-specific fine-tuning, which can involve substantial computational resources and memory. Furthermore, models predominantly pre-trained on high-resource languages may struggle with domain-specific expressions and linguistic variations in low-resource languages such as Roman Urdu. These challenges highlight the importance of efficient adaptation strategies that reduce training requirements while maintaining classification performance [12]

Recent advancements in large language models (LLMs), including GPT, LLaMA, Mistral, and Gemma, have introduced new opportunities for hate speech detection. These models support zero-shot and few-shot learning, allowing classification through natural language instructions without necessarily updating model parameters [3]. Prompt engineering enables task adaptation through carefully designed instructions, while in-context learning incorporates labeled examples within prompts to improve classification. Such approaches are particularly valuable in low-resource settings where extensive annotated datasets and computational infrastructure may be unavailable. Recent studies have explored both prompt-based learning and efficient fine-tuning for text classification. Xu et al. [13] investigated QLoRA-based adaptation of LLaMA2-7B for news topic classification, reporting improvements over selected transformer baselines. Nadeau et al. [14] evaluated LLMs, including LLaMA2, Mistral, Gemma, and GPT, across safety-related dimensions such as toxicity, bias, factuality, and hallucination, highlighting differences in model behavior. These findings demonstrate the importance of selecting appropriate models and adaptation strategies for reliable classification.

Prompt-based techniques have also shown promising results in hate speech detection. Zhang et al. [15] introduced Domain-Enhanced Prompt Learning (DePL) for implicit hate speech detection in Chinese, addressing data scarcity through domain-aware prompt learning. Yamagishi [16] explored simplified prompts for zero-shot multimodal hate speech detection, suggesting that concise prompts can be effective without additional complexity. Similarly, Govers et al. [17] proposed Prompt-GAN, employing adversarial prompt tuning to generate synthetic extremist content and reporting improvements in classification performance. Han and Tang [18] investigated GPT-3-based in-context learning for hate speech detection, emphasizing the influence of task instructions and demonstration examples on few-shot classification.

Although these studies demonstrate the potential of LLMs and prompt-based learning, their effectiveness remains sensitive to prompt structure, demonstration selection, and linguistic characteristics. These limitations are particularly relevant to Roman Urdu, where inconsistent spelling, code-mixing, informal expressions, and contextual ambiguity complicate classification. Consequently, systematic evaluation of zero-shot and few-shot prompting, together with alternative prompt tuning strategies, is necessary to identify effective approaches for low-resource hate speech detection.

Despite significant advancements in hate speech detection, several research gaps remain, particularly for low-resource languages such as Roman Urdu. Existing studies predominantly focus on high-resource languages, with limited attention to Roman Urdu's linguistic complexities, including inconsistent spelling, code-mixing, and contextual ambiguity. Furthermore, comprehensive evaluations of zero-shot and few-shot prompting, manual and mixed prompt tuning, and their comparative effectiveness remain limited. Although parameter-efficient fine-tuning (PEFT) has demonstrated promising results, further investigation is needed to understand how prompt-based approaches compare with previously reported PEFT benchmarks. Additionally, the relationship between classification performance, computational efficiency, and adaptation complexity remains insufficiently explored. This study addresses these gaps by systematically evaluating prompt engineering and prompt tuning strategies for Roman Urdu hate speech detection, aiming to identify effective and resource-efficient approaches for low-resource NLP applications.

## III. Material and Methods

### *3.1. Dataset Details*

The PURUTT (Parallel Urdu and Roman Urdu Toxicity) dataset [2] is a benchmark for hate speech detection in low-resource languages, comprising 72,771 social media comments, including 13,097 toxic and 59,674 non-toxic samples. It reflects real-world linguistic challenges such as inconsistent spelling, phonetic variations, code-mixing, informal grammar, and semantic ambiguity, making it suitable for evaluating NLP models. For experimentation, the dataset is preprocessed and divided into 60% training, 20% validation, and 20% testing subsets. To address class imbalance, class-weighted cost-sensitive learning is applied during training. Its linguistic diversity and real-world relevance make PURUTT an effective benchmark for evaluating large language models, prompt-based learning, and parameter-efficient fine-tuning techniques such as LoRA. Summary of PURUTT dataset is described in Table 1.

TABLE 1. SUMMARY OF PURUTT DATASET

| Attribute | Description |
|---|---|
| Total Samples | 72,771 Nos. |
| Toxic Samples | 13,097 Nos. |
| Non-Toxic Samples | 59,674 Nos. |
| Task Type | Binary Classification (Toxic / Non-Toxic) |
| Language | Roman Urdu (Latin script) |
| Data Source | Social media comments |
| Class Distribution | Imbalanced (≈18% toxic, 82% non-toxic) |
| Key Challenges | Non-standard spelling, phonetic variation, semantic ambiguity |
| Preprocessing | Tokenization, normalization, padding/ truncation |
| Data Split | 60% Train / 20% Validation / 20% Test |
| Handling Imbalance | Class weighting (cost-sensitive learning) |

### *3.2. Proposed Methodology*

This study presents a unified framework for evaluating prompt engineering and prompt tuning for Roman Urdu hate speech detection. The framework addresses key challenges of low-resource NLP, including inconsistent spelling, code-mixing, semantic ambiguity, and class imbalance. Unlike conventional fine-tuning, the proposed approaches leverage pre-trained language models while minimizing the need for extensive parameter updates. ***Fig. 1*** illustrates the proposed architecture. The process begins with the PURUTT dataset, followed by text normalization, tokenization, padding, and truncation. The framework investigates three complementary adaptation strategies. First, **Prompt engineering** evaluates zero-shot and few-shot classification using GPT-3.5 through Azure OpenAI, employing task-specific instructions and labeled examples without updating model parameters.

Second, **Prompt tuning** utilizes multilingual BERT (uncased) through the OpenPrompt framework to evaluate Prefix, Cloze, and Mixed Prompting strategies under few-shot configurations of (K = 32,64,128). The Mixed Prompting approach combines fixed textual templates with learnable soft embeddings to improve task-specific representations. Previously published results [19] obtained using conventional inference and PEFT-LoRA are considered as reference baselines for contextual comparison rather than as experimental contributions of the present study. The proposed approaches are evaluated using accuracy, precision, recall, and F1 score to assess their effectiveness for Roman Urdu hate speech detection.

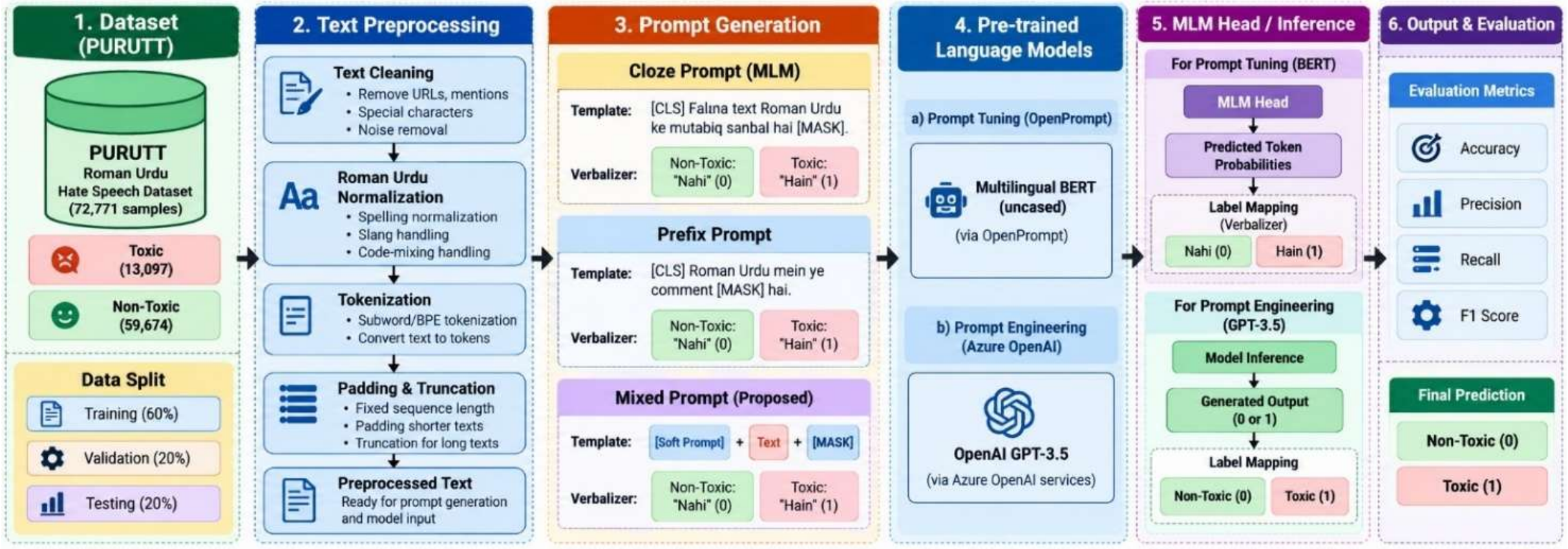

**Fig. 1.** Proposed framework for Roman Urdu Hate Speech Detection (RU-HSD) using prompt engineering and prompt tuning strategies.

### *3.2.1. Prompt Engineering*

Prompt engineering is employed to evaluate the ability of large language models to classify Roman Urdu comments without updating model parameters. The classification task is formulated through structured chat messages consisting of system instructions, user inputs, and model-generated responses. Two strategies are investigated: ***zero-shot and few-shot prompting.***

#### *A. Zero Shot Prompting:*

In zero-shot prompting, the model receives a task-specific system instruction and an unlabeled Roman Urdu comment. No labeled examples are provided. The model is instructed to classify the comment as *toxic (1)* or *non-toxic (0)*, returning only the predicted label. The following system prompt is employed:

**Prompt:**

You are a hate speech detection assistant. You will be provided with a comment. Each comment will be enclosed by triple backticks. You need to read the comment and classify it into one of the below two categories:

- 1 - for toxic comments (containing hate speech, offensive language, or harmful content).
- 0 - for non-toxic comments (free of hate speech and offensive language).

Do NOT use any other categories like Neutral or Mixed. If both types of content exist, choose the dominant tone. This is your FINAL CLASSIFICATION. Only output 0 or 1. No explanations.

The zero-shot approach provides a baseline for evaluating the inherent classification capabilities of pre-trained LLMs in a linguistically complex, low-resource environment.

#### *B. Few Shot Prompting:*

Few-shot prompting extends the zero-shot approach by introducing labeled examples before the target comment. The thesis uses five toxic and five non-toxic examples in each prompt, represented as user–assistant message pairs. These examples provide contextual guidance without modifying the model parameters. The same system instruction is retained to ensure consistent classification criteria. The research evaluates GPT-3.5 through Microsoft Azure OpenAI, processing comments in batches of 100. It reports 728 batches for each prompting configuration, with batch-level evaluation metrics averaged to obtain the final results.

**Algorithm 1: Prompt Engineering using OpenAI for RU-HSD**

**Input:** D: Roman Urdu dataset with comments and labels
**Input:** M: OpenAI GPT-3.5 model
**Output:** Predicted labels and evaluation metrics

Define system message with strict instructions to output only 0 or 1
Split dataset D into batches of 100
**for each batch $B_i$ in D do**
    Initialize prompt with system message
    **if few-shot then**
        Select 5 examples per class from $B_i$ and add to prompt
    **end if**
    **for each comment c in $B_i$ do**
        Add user message with c
        Model generates predicted label (0 or 1)
        Save predicted label
    **end for**
    Compute batch-level metrics: accuracy, precision, recall, F1-score
**end for**
Compute final overall metrics as the average of all batch metrics
Return labeled dataset and overall metrics

### *3.2.2. Prompt Tuning*

Prompt tuning is investigated using the ***OpenPrompt framework*** and the *bert-base-multilingual-uncased model*. Roman Urdu hate speech detection is reformulated as a masked language modeling task, where a [MASK] token is used to predict label-related words. A manual verbalizer maps these predictions to toxic and non-toxic classes. The experiments consider manual prefix and cloze templates alongside a mixed prompt approach incorporating learnable soft tokens.

#### *A. Manual Prompt:*

Manual prompting employs fixed, human-designed templates to formulate hate speech detection as a masked-token prediction task. Two template structures are evaluated:

| | |
|---|---|
| **Prefix Prompt:** | **Ye hai [MASK]!** |
| **Cloze Prompt:** | **Ye [MASK] hai.** |

In the prefix formulation, the masked token appears at the end of the template, whereas the cloze formulation places it within the sentence. The input comment provides contextual information for predicting the masked token, which is subsequently mapped to a classification label through the verbalizer. Both formulations use fixed textual templates without introducing learnable soft prompt embeddings. Their comparison enables an assessment of the influence of template structure and masked-token placement on classification performance.

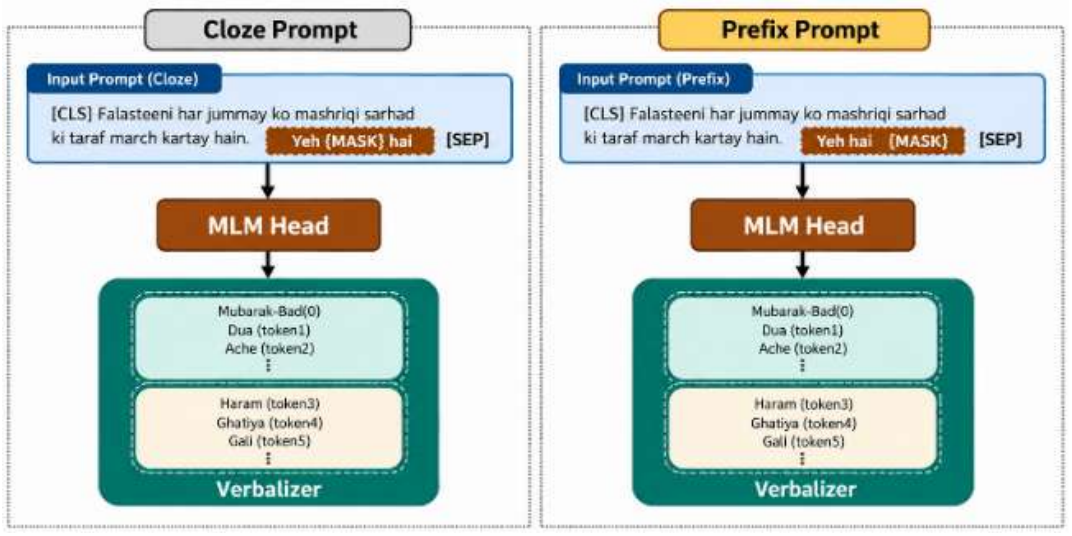

**Fig. 2.** Architecture of Prefix-Based and Cloze-Based Manual Prompting

## *B. Mixed Prompt:*

Mixed prompt tuning combines manually constructed prompt elements with trainable soft token embeddings. Unlike fixed manual prompts, the soft embeddings are optimized during training while the underlying language model remains frozen. This allows task-specific adaptation with a limited number of trainable parameters. The mixed prompt format is as follows:

**{"placeholder":"text_a","soft":"Yeh","soft":"comment hai","mask"}**

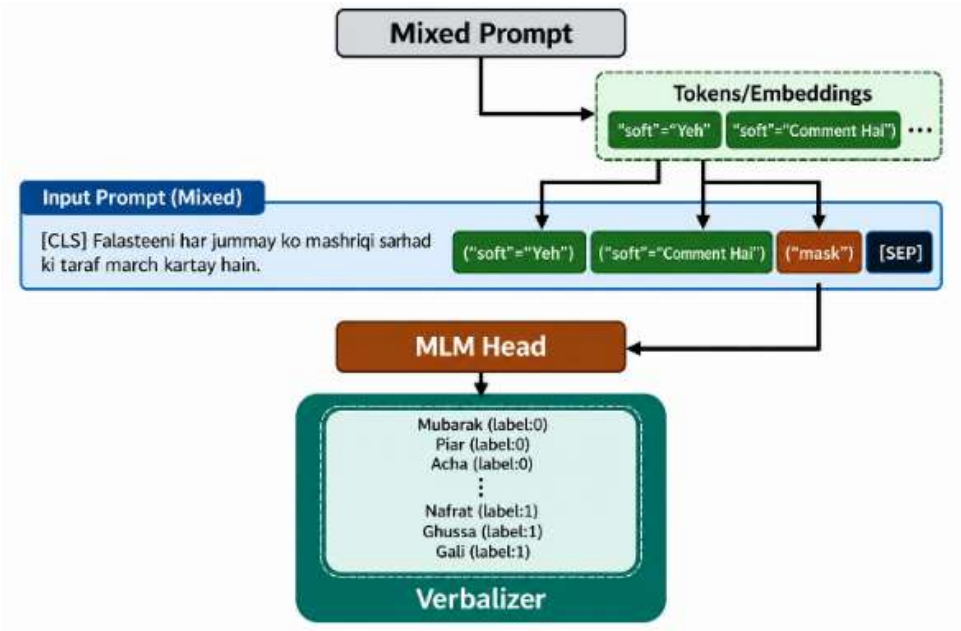

**Fig. 3.** Mixed Prompt Tuning Architecture with Learnable Soft Tokens

This hybrid approach combines human-defined prompt structure with data-driven adaptation. It aims to improve the representation of linguistic variations and contextual information while reducing the number of trainable parameters compared with full-model fine-tuning.

## *C. Manual Verbalizer*

A manual verbalizer converts masked-token predictions into binary class labels. The thesis defines label-word groups associated with non-toxic (0) and toxic (1) comments. The verbalizer enables the model to associate predicted words with the corresponding classification labels.

TABLE 2. MANUAL VERBALIZER FOR RU-HSD

| Class | Label words |
|---|---|
| Non-Toxic (0) | muhabbat, pyaar, acha, dua, wah |
| Toxic (1) | khilaf, nafrat, ghussa, lanat, ganda, beh***d, gaali, bura |

**Algorithm 1. Prompt Tuning Using the OpenPrompt Framework**

**Input:** D: Roman Urdu dataset containing comments and labels

**Input:** PLM: Pretrained language model (bert-base-multilingual-uncased)

**Input:** T: Prompt template style (prefix, cloze, or mixed)

**Output:** Trained prompt-based classifier

```
Split D into D_train+dev (80%) and D_test (20%)
Define a verbalizer mapping label words to classes 0 and 1
for sample size n in {32, 64, 128} do
    for repetition r = 1 to 2 do
        Randomly sample n examples from D_train+dev as D_train^(r)
        Use the remaining examples as D_dev^(r)
        Initialize the OpenPrompt pipeline with PLM
        if T = prefix then
            Template ← "Ye hai [MASK]!"
        else if T = cloze then
            Template ← "Ye [MASK] hai."
        else if T = mixed then
            Template ← {"placeholder":"text_a", "soft":"Yeh", "soft":"comment hai", "mask"}
        end if
        Create PromptDataLoader for D_train^(r)
        Initialize loss and optimizer (for soft tokens if T = mixed)
        for epoch = 1 to 3 do
            for each batch in D_train^(r) do
                Compute loss and update parameters
            end for
        end for
        Evaluate on D_dev^(r) and store results
    end for
    Select the best model among both repetitions
    Evaluate the selected model on D_test
end for
```

## *3.3. Implementation Details*

- The proposed framework is implemented in Python using PyTorch, Hugging Face Transformers, and OpenPrompt. Prompt engineering experiments are conducted using GPT-3.5 through Azure OpenAI services, while prompt tuning employs multilingual BERT (uncased) with fixed textual templates and learnable soft prompt representations. Model-specific tokenization, padding, and truncation are applied to ensure appropriate input processing. The local prompt-tuning experiments are conducted using an NVIDIA A100 GPU. For prompt tuning, AdamW optimization is employed with a learning rate of ***1 x 10⁻⁴***, batch size of ***32, 64, and 128***, and ***3 training epochs***. The maximum sequence length is ***512 tokens***, with ***dropout set to 0.05***. Validation-based model selection is used before final evaluation on the test set. The main experimental configurations are summarized below.

TABLE 3. EXPERIMENTAL CONFIGURATION

| Component | Specification |
|---|---|
| Framework | PyTorch, |
| Prompt Engineering Model | GPT 3.5 (Azure Open AI) |
| Prompt Tuning Model | BERT-base-multilingual-uncased |
| GPU | NVIDIA A100 |
| Sequence Length | 512 tokens |
| Batch Size | 8 |
| Optimizer | AdamW |
| Learning Rate | $1 \times 10^{-4}$ |
| Epochs | 3 |
| Dropout | 0.05 |
| Prompt Engineering | Zero-shot, Few-shot |
| Prompt Tuning | Manual, Mixed, Cloze |
| Evaluation | Accuracy, Precision, Recall, F1-score |

## IV. Results and Discussion

This section presents the experimental results of prompt engineering and prompt tuning for Roman Urdu hate speech detection using the PURUTT dataset. The evaluation examines the effectiveness of zero-shot and few-shot prompting, manual prompt formulations, and mixed prompt tuning across different pre-trained language models. The primary objective is to investigate how these adaptation strategies influence classification performance in a low-resource language characterized by spelling variations, code-mixing, informal expressions, and class imbalance. Model performance is evaluated using accuracy, precision, recall, and F1-score. Particular emphasis is placed on F1-score because the dataset contains substantially fewer toxic samples than non-toxic samples. The experimental analysis is organized into three parts: performance evaluation of prompt engineering, comparison of manual and mixed prompt tuning, and model-wise comparative analysis. Previously reported parameter-efficient fine-tuning (PEFT) results are considered separately to provide additional context regarding the effectiveness of prompt-based adaptation.

### *4.1. Performance Evaluation of Prompt Engineering*

#### *4.1.1. Experimental Approach*

This experiment evaluates the effectiveness of zero-shot and few-shot prompt engineering for Roman Urdu hate speech detection using the OpenAI GPT-3.5 model accessed through Azure OpenAI services. The objective was to investigate the model's classification performance without task-specific fine-tuning under two prompting configurations:

- **Zero-Shot Prompting:** The model was provided with task-specific instructions without labeled examples.
- **Few-Shot Prompting:** A small set of representative labeled examples was included in the prompt to guide the model's classification decisions.

For both configurations, the Roman Urdu comments were processed in batches of 100. Performance was evaluated using accuracy, precision, recall, and F1 score. The reported results include averaged evaluation metrics and the maximum F1 score observed across the evaluated batches.

#### *4.1.2. Experimental Results:*

Table 4 presents the comparative performance of zero-shot and few-shot prompting. The results demonstrate that few-shot prompting consistently outperformed zero-shot prompting across all four-evaluation metrics.

Table 4. Performance Comparison of Zero-Shot and Few-Shot Prompt Engineering

| Prompt Type | Accuracy | Precision | Recall | F1-Score | Max F1-Score |
|---|---|---|---|---|---|
| Few-Shot | **0.9100** | **0.8278** | **0.8989** | **0.8565** | **0.9494** |
| Zero-Shot | 0.7229 | 0.6799 | 0.8129 | 0.6698 | 0.8250 |

Few-shot prompting achieved an accuracy of 0.9100 and F1 score of 0.8565, compared with 0.7229 and 0.6698 for zero-shot prompting. This represents absolute improvements of 18.71 and 18.67 percentage points in accuracy and F1 score, respectively. Similarly, precision increased from 0.6799 to 0.8278, while recall improved from 0.8129 to 0.8989. The maximum observed F1 score also increased from 0.8250 for zero-shot prompting to 0.9494 for few-shot prompting. Figure 4.11 illustrates these performance differences, highlighting the consistent advantage of providing labeled examples during inference.

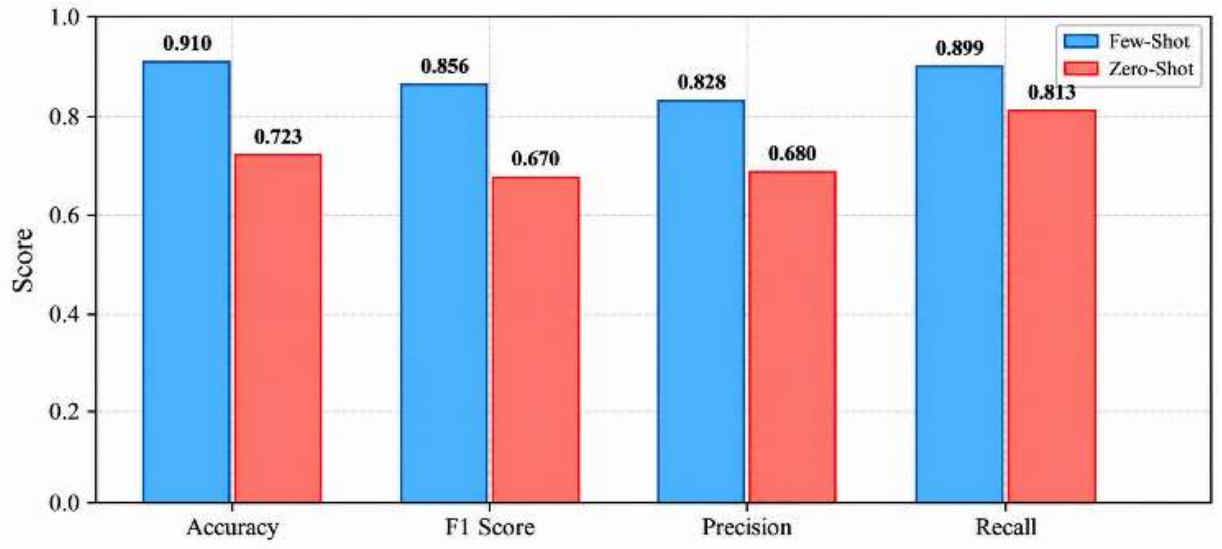

**Fig. 4.** Comparative performance of GPT-3.5 using zero-shot and few-shot prompting across accuracy, precision, recall, and F1 score.

### *4.2. Performance Evaluation of Prompt Tuning*

#### *4.2.1. Experimental Approach*

This experiment evaluates multilingual BERT (uncased) for Roman Urdu hate speech detection using three adaptation strategies: direct inference without fine-tuning, Parameter-Efficient Fine-Tuning (PEFT) using Low-Rank Adaptation (LoRA), and prompt tuning. Three prompting strategies were evaluated: **Prefix Prompting, Cloze Prompting, and Mixed Prompting**. Each strategy was assessed using three values of the prompt configuration parameter***, (K = 32, 64, 128).*** The performance was measured using precision, recall, F1 score, and classification accuracy.

#### *4.2.2. Dataset Preparation and Experimental Setup*

The PURUTT dataset was divided into three mutually exclusive subsets: ***training*** $(D_{train})$, development $(D_{dev})$, and testing $(D_{test})$, satisfying:

$$D = D_{train} \cup D_{dev} \cup D_{test} \quad (1)$$

$$D_{train} \cap D_{dev} = D_{train} \cap D_{test} = D_{dev} \cap D_{test} = \emptyset \quad (2)$$

For each few-shot configuration, K training examples and K development examples were randomly selected, where:

$$K \in \{32, 64, 128\} \quad (3)$$

$$|Dtrain| = K, \quad |Ddev| = K, \quad |Dtest| = |D| - 2K \quad (4)$$

Each configuration was evaluated over two experimental rounds, with two epochs per round:

$$R = 2, \quad E = 2 \quad (5)$$

Training samples were selected without overlap between experimental rounds:

$$D_{train^{(i)}} \cap D_{train^{(j)}} = \emptyset, \quad i \neq j \quad (6)$$

The same preprocessing and evaluation procedures were applied across the experiments to support consistent performance comparisons.

#### *4.2.3. Evaluation of Prefix Prompting*

Table 5 and Fig. 5presents the performance of Prefix Prompting across different values of ***(K = 32, 64, and 128).*** At (K=32), the model achieved an F1 score of 0.6095 and accuracy of 0.7970. Increasing (K) to 64 improved the F1 score to 0.6312, although accuracy decreased to 0.7584. The best performance was obtained at (K=128), achieving a precision of 0.6747, recall of 0.6894, F1 score of 0.6813, and accuracy of 0.8048. These results indicate that increasing (K) generally improved the F1 score, although accuracy did not follow a consistent trend.

TABLE 5. EVALUATION OF PREFIX PROMPTING BY DIFFERENT K-VALUES

| K-Value | Precision | Recall | F1-Score | Accuracy |
|---|---|---|---|---|
| 32 | 0.6311 | 0.5989 | 0.6095 | 0.7970 |
| 64 | 0.6222 | 0.6485 | 0.6312 | 0.7584 |
| 128 | 0.6747 | 0.6894 | 0.6813 | 0.8048 |

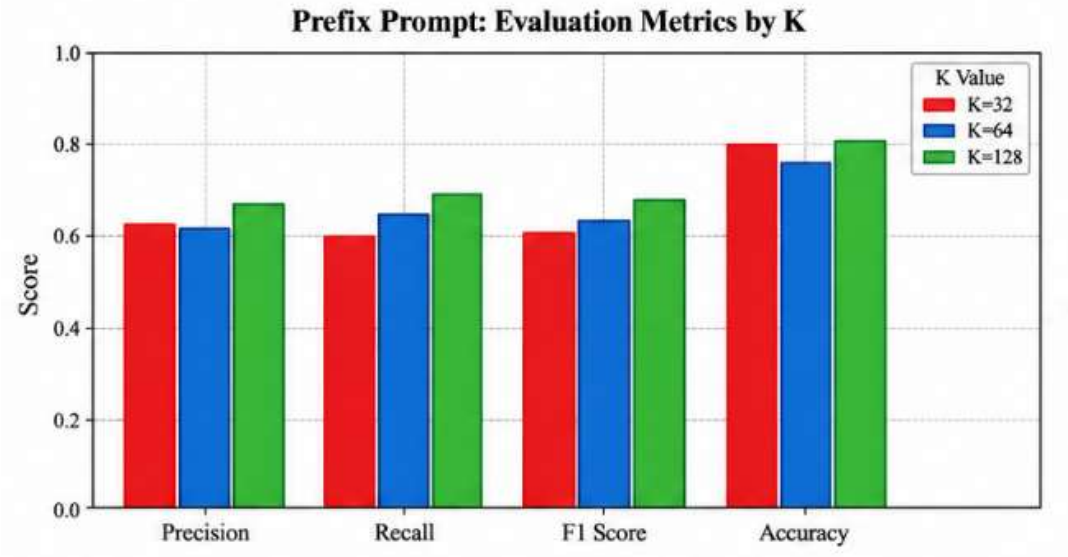

**Fig. 5.** Prefix Prompt: Evaluation Metrics by K

### 4.2.4. *Evaluation of Cloze Prompting*

Table 6 and Fig. 6 summarizes the results of Cloze Prompting. At (K=32), the model achieved its highest F1 score of 0.6532 and accuracy of 0.6984. Increasing (K) to 64 improved recall to 0.7234 but reduced the F1 score to 0.5886 and accuracy to 0.6299. At (K=128), the F1 score recovered to 0.6167, while accuracy reached 0.6842. These findings suggest that increasing (K) does not consistently improve Cloze Prompting performance, with (K=32) providing the best balance between F1 score and accuracy.

TABLE 6. EVALUATION OF CLOZE PROMPTING BY DIFFERENT K-VALUES

| K-Value | Precision | Recall | F1-Score | Accuracy |
|---|---|---|---|---|
| 32 | 0.6097 | 0.6833 | 0.6532 | 0.6984 |
| 64 | 0.6342 | 0.7234 | 0.5886 | 0.6299 |
| 128 | 0.6256 | 0.7049 | 0.6167 | 0.6842 |

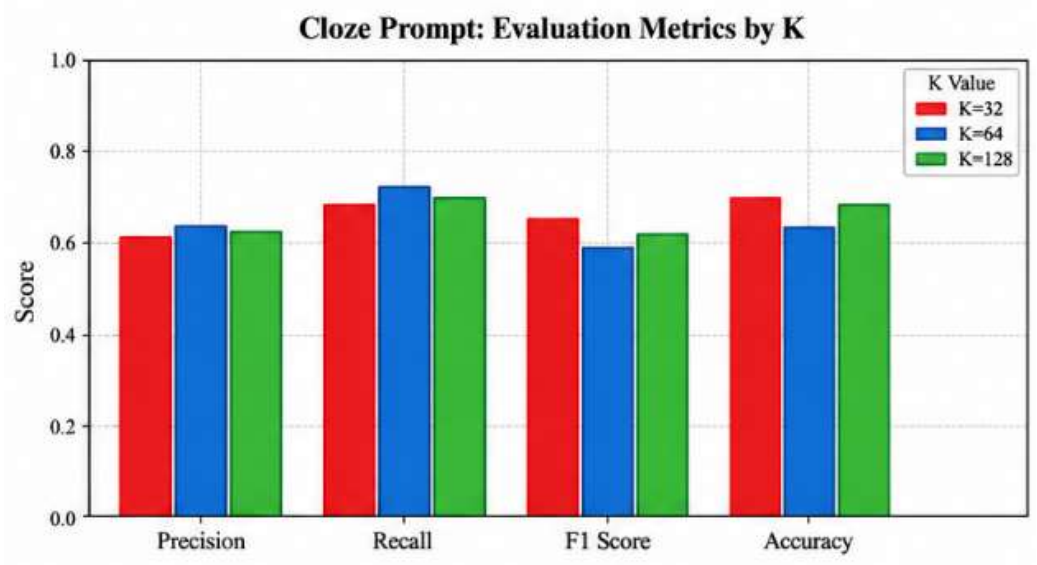

**Fig. 6.** Cloze Prompt: Evaluation Metrics by K

### 4.2.5. *Evaluation of Mixed Prompting*

Table 8 and Fig. 7 presents the results of Mixed Prompting, which combines fixed textual prompts with learnable prompt representations. At (K=32), the model achieved an F1 score of 0.6008 and accuracy of 0.6976. Increasing (K) to 64 improved the F1 score to 0.6531 and accuracy to 0.7580. At (K=128), the model achieved its highest F1 score of 0.6852 and recall of 0.7810, although accuracy decreased slightly to 0.7498. Overall, Mixed Prompting demonstrated improved F1 performance with increasing (K), achieving the highest recall among all evaluated prompting configurations.

TABLE 7. EVALUATION OF MIXED PROMPTING BY DIFFERENT K-VALUES

| K-Value | Precision | Recall | F1-Score | Accuracy |
|---|---|---|---|---|
| 32 | 0.5992 | 0.6484 | 0.6008 | 0.6976 |
| 64 | 0.6405 | 0.6885 | 0.6531 | 0.7580 |
| 128 | 0.6774 | 0.7810 | 0.6852 | 0.7498 |

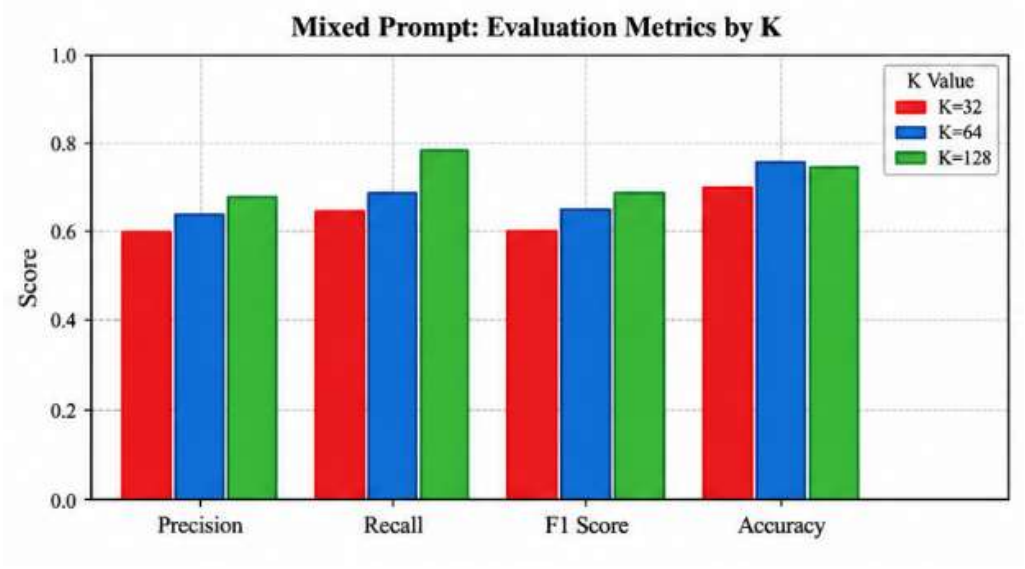

**Fig. 7.** Mixed Prompt: Evaluation Metrics by K

### 4.2.6. *Comparative Analysis of Prompting Strategies*

The comparative results in Table 8 demonstrate that Mixed Prompting achieved the highest F1 score of 0.6852 at (K=128), followed closely by Prefix Prompting (0.6813). Prefix Prompting achieved the highest classification accuracy of 0.8048, while Cloze Prompting achieved its best F1 score of 0.6532 at (K=32). These findings indicate that Mixed Prompting was more effective in terms of F1 score and recall, whereas Prefix Prompting provided better overall accuracy.

TABLE 8. ANALYSIS OF PERFORMANCE ACROSS PROMPTING STRATEGIES

| Prompt Type | Best F1-Score | Best Accuracy |
|---|---|---|
| Prefix | 0.6813 | 0.8048 |
| Cloze | 0.6532 | 0.6984 |
| Mixed | 0.6852 | 0.7580 |

The results in Table 9 and Fig. 8 highlight the effectiveness of prompt-based approaches for Roman Urdu hate speech detection. Among previously published baselines [19], Mistral achieved the highest F1 score (0.9387) and accuracy (0.9642). In the current study, GPT-3.5 few-shot prompting outperformed zero-shot prompting, achieving an F1 score of 0.8565 and accuracy of 0.9100, demonstrating the benefits of incorporating labelled examples without fine-tuning.

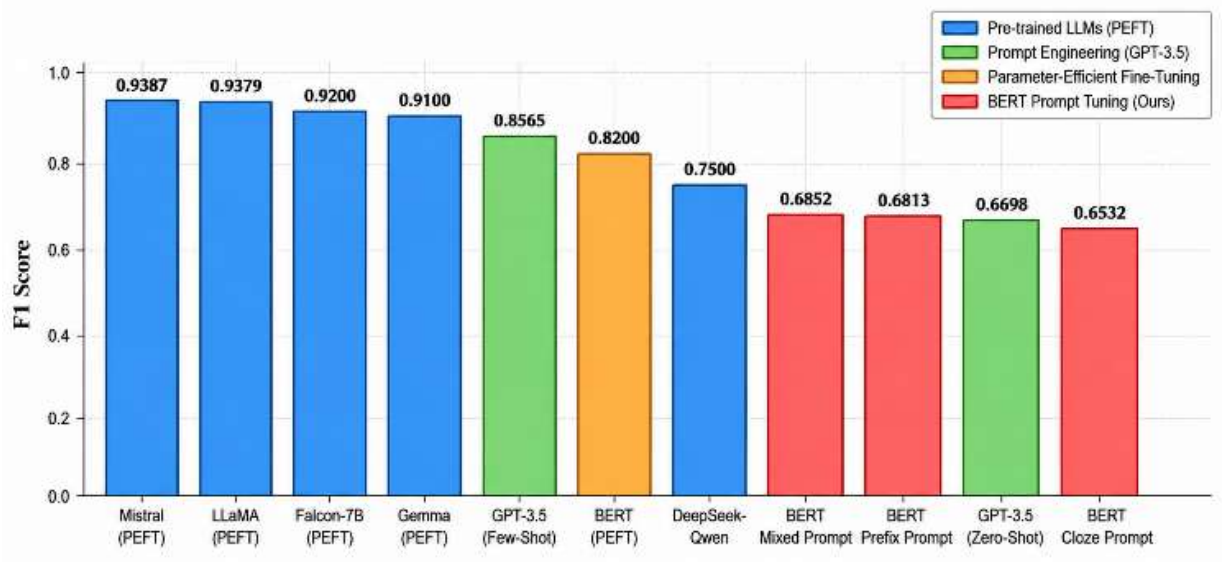

**Fig. 8.** F1 score comparison of prompting strategies and PEFT-based models for Roman Urdu hate speech detection.

TABLE 9. PERFORMANCE COMPARISON OF PROMPTING STRATEGIES AND PEFT-BASED MODELS FOR ROMAN URDU HATE SPEECH DETECTION

| Technique | F1 | Acc. | Prec. | Recall |
|---|---|---|---|---|
| Mistral (PEFT) [19] | 0.9387 | 0.9642 | 0.9383 | 0.9392 |
| LLaMA (PEFT) [19] | 0.9379 | 0.9630 | 0.9407 | 0.9353 |
| Falcon-7B (PEFT) [19] | 0.9200 | 0.9500 | 0.9300 | 0.9100 |
| Gemma (PEFT) [19] | 0.9100 | 0.9463 | 0.9100 | 0.9100 |
| **GPT-3.5 (Few-Shot)** | **0.8565** | **0.9100** | **0.8278** | **0.8989** |
| BERT (PEFT) [19] | 0.8200 | 0.8945 | 0.8200 | 0.8200 |
| DeepSeek-Qwen | 0.7500 | 0.9354 | 0.7300 | 0.7800 |
| BERT (Mixed Prompt) | 0.6852 | 0.7498 | 0.6774 | 0.7810 |
| BERT (Prefix Prompt) | 0.6813 | 0.8048 | 0.6747 | 0.6894 |
| GPT-3.5 (Zero-Shot) | 0.6698 | 0.7229 | 0.6799 | 0.8129 |
| BERT (Cloze Prompt) | 0.6532 | 0.6984 | 0.6097 | 0.6833 |

Among BERT-based strategies, the proposed Mixed Prompt Tuning achieved the highest F1 score (0.6852) and recall (0.7810) at (K=128), while Prefix Prompting achieved the highest accuracy (0.8048). Cloze Prompting obtained its best F1 score (0.6532) at (K=32). Increasing few-shot examples generally improved Prefix and Mixed Prompting performance. Previously published PEFT-LoRA results (F1: 0.8200; accuracy: 0.8945) exceeded the BERT-based prompting results, although experimental differences limit direct comparisons. Overall, the findings demonstrate the potential of few-shot prompt engineering and Mixed Prompt Tuning for low-resource hate speech detection, highlighting the importance of prompt design and labelled examples.

## V. Conslusions

This study investigated the effectiveness of prompt engineering and prompt tuning for RU-HSD using the PURUTT dataset. The experimental evaluation demonstrated that GPT-3.5 few-shot prompting outperformed zero-shot prompting, achieving an F1 score of 0.8565 and accuracy of 0.9100, highlighting the importance of representative labeled examples in low-resource language classification. Among the BERT-based prompting strategies, the proposed Mixed Prompt Tuning approach, combining fixed textual templates with learnable soft embeddings, achieved the highest F1 score of 0.6852 and recall of 0.7810, while Prefix Prompting achieved the highest accuracy of 0.8048. The results further revealed that increasing few-shot training examples improved Prefix and Mixed Prompting performance, although improvements were not consistent across all configurations. Comparisons with previously published PEFT-LoRA [19] and language model baselines provided additional context for evaluating the proposed approaches. Overall, the findings demonstrate the potential of prompt-based learning for Roman Urdu hate speech detection without extensive model fine-tuning. Future research will focus on improving prompt optimization, investigating cross-domain generalization, and evaluating the proposed approaches across additional low-resource languages and datasets.